\documentclass{ecai}
\usepackage{graphicx}
\usepackage{latexsym}

\ecaisubmission   

\paperid{527}     

\usepackage{times}
\usepackage{url}
\usepackage[hidelinks]{hyperref}
\usepackage[utf8]{inputenc}
\usepackage[small]{caption}
\usepackage{graphicx}
\usepackage{amsmath}
\usepackage{amsthm}
\usepackage{algorithm}
\usepackage{algorithmic}

\usepackage{multirow}
\usepackage{subfigure}
\usepackage{newtxmath}
\usepackage{makecell}
\DeclareMathOperator{\st}{s.t.}

\usepackage{latexsym}

\begin{document}
	
	\begin{frontmatter}
		
		\title{OER: Offline Experience Replay for Continual Offline Reinforcement Learning}
		
		\author[AB]{\fnms{Sibo}~\snm{Gai}\thanks{Email: gaisibo@westlake.edu.cn}}
		\author[B]{\fnms{Donglin}~\snm{Wang}\thanks{Corresponding Author. Email: wangdonglin@westlake.edu.cn}}
		\author[B]{\fnms{Li}~\snm{He}} 

		\address[A]{Fudan University, Shanghai, China}
		\address[B]{Westlake University, Zhejiang, China}
		
		\begin{abstract}
			The capability of continuously learning new skills via a sequence of pre-collected offline datasets is desired for an agent. 
			However, consecutively learning a sequence of offline tasks likely leads to the catastrophic forgetting issue under resource-limited scenarios. 
			In this paper, we formulate a new setting, continual offline reinforcement learning (CORL), where an agent learns a sequence of offline reinforcement learning tasks and pursues good performance on all learned tasks with a small replay buffer without exploring any of the environments of all the sequential tasks.
			For consistently learning on all sequential tasks, an agent requires acquiring new knowledge and meanwhile preserving old knowledge in an offline manner.
			To this end, we introduced continual learning algorithms and experimentally found experience replay (ER) to be the most suitable algorithm for the CORL problem. However, we observe that introducing ER into CORL encounters a new distribution shift problem: the mismatch between the experiences in the replay buffer and trajectories from the learned policy. 
			To address such an issue, we propose a new model-based experience selection (MBES) scheme to build the replay buffer, where a transition model is learned to approximate the state distribution. This model is used to bridge the distribution bias between the replay buffer and the learned model by filtering the data from offline data that most closely resembles the learned model for storage.
			Moreover, in order to enhance the ability on learning new tasks, we retrofit the experience replay method with a new dual behavior cloning (DBC) architecture to avoid the disturbance of behavior-cloning loss on the Q-learning process.
			In general, we call our algorithm offline experience replay (OER).
			Extensive experiments demonstrate that our OER method outperforms SOTA baselines in widely-used Mujoco environments.
			
		\end{abstract}
	\end{frontmatter}
	
	\section{Introduction}
	Similar to human beings, a general-purpose intelligence agent is expected to learn new tasks continually. 
	Such sequential tasks can be either online tasks learned through exploration or offline tasks learned through offline datasets, where the latter is equally important but has not drawn sufficient attention so far.
	Learning sequential tasks in offline setting can greatly improve the learning efficiency and avoid the dangerous exploration process in practice.
	Moreover, online learning sequential tasks is not always feasible due to the temporal and spatial constraints of the environment and the agent’s on-board consideration itself.  
	Therefore, studying offline reinforcement learning in the continual setting is quite important and valuable for general-purpose intelligence.
	If having sufficient computational resources, it is easy to accomplish such a goal. 
	However, for an agent with limited resources \footnote{Running on a server and communicating with the agent are undesirable due to complex-system, real-time, and data privacy issues.}, continual learning methods are indispensable to deal with such offline datasets.
	Consequently, we propose a new setting named continual offline reinforcement learning (CORL) in this paper, which integrates offline RL and continual learning.
	
	Offline RL learns from pre-collected datasets instead of directly interacting with the environment \cite{BCQ}.
	By leveraging pre-collected datasets, offline RL methods avoid costly interactions and thus enhance learning efficiency and safety. 
	However, offline RL methods suffer from the over-estimation problem of the out-of-distribution (OOD) data, where the unseen data are erroneously estimated to be high values.
	This phenomenon stems from the distribution shift between the behavior policy and the learning policy. 
	Various methods have been proposed to address the over-estimation problem
	\cite{BCQ,BEAR}.
	In this paper, we focus on dealing with a sequence of offline datasets, which requires continual learning techniques.
	
	The major challenge of continual learning is how to alleviate the catastrophic forgetting \cite{CataInter} issue on previous tasks when learning new tasks. 
	There are three types of continual learning methods, including regularization-based methods \cite{EWC,SI}, modular methods \cite{PathNet,PackNet}, and rehearsal-based methods \cite{GEM,AGEM}.
	Experience replay (ER) is a widely-used rehearsal-based method \cite{ER}, which alternates between learning a new task and replaying samples of previous tasks. 
	In this paper, we consider using ER as a base for our own problem. We will show that ER is the most suitable algorithm for the CORL problem in the experiment section following.
	
	However, since ER is designed for online continual RL, directly applying ER in our CORL yields poor performance due to two kinds of distribution shifts. 
	The first is the distribution shift between the behavior policy and the learning policy, and the second is the distribution shift between the selected replay buffer and the corresponding learned policy. 
	Existing methods focus on addressing the first shift issue, and no related work considers the second, which only appears in our CORL setting.
	Thus, simply integrating ER with Offline RL fails to alleviate the catastrophic forgetting.
	To solve the new problem above, we propose a novel model-based experience selection (MBES) method to fill the replay buffer.
	The key idea is to take advantage of the dynamic model to search for and add the most valuable episodes in the offline dataset into the replay buffer.
	
	After having a good replay buffer for previous tasks, the learned policy corresponding to a new task needs to clone previous tasks. 
	Behavior cloning (BC) as an online ER method is widely used  \cite{HyperFree}, which is incompatible with the actor-critic architecture in offline RL.
	Even though we can carefully tune the weight of BC loss, tuning the hyper-parameter is a cumbersome task in general and is difficult in the offline setting.
	Therefore, integrating online ER with offline RL often derives a non-convergent policy.
	The reason is that the actor-critic model is hard to train \cite{GANAC}, and the rehearsal term in the loss function has a negative effect on the learning process. 
	To effectively replay experiences, we propose a dual behavior cloning (DBC) architecture instead to resolve the optimization conflict, where one policy optimizes the performance of the new task by using actor-critic architecture, and the second optimizes from the continual perspective for both new and learned tasks.

	In summary, this paper considers investigating a new setting CORL.
	A novel MBES is proposed to select valuable experiences and overcome the mismatch between the experiences in the replay buffer and trajectories from the learned policy. 
	Then, a DBC architecture is proposed to deal with the optimization conflict problem. 
	By taking MBES and DBC as two key ideas for CORL setting, we name our overall scheme as offline experience replay (OER). 
	The main contributions of this paper can be summarized as follows:
	\begin{itemize}
		\item We present a new setting CORL and then propose a novel scheme OER for CORL setting.
		\item We propose a novel selection method MBES for offline replay buffer by utilizing a dynamic model to reduce the distribution shift between experiences from replay buffer and learned policy. 
		\item On the other hand, we propose a novel DBC architecture to prevent the learning process from collapsing by separating the Q-learning on current task and BC processes on all previous tasks.
		\item We experimentally verify the performance of different modules and evaluate our method on continuous control tasks. Our method OER outperforms all SOTA baselines for all cases.
	\end{itemize}
	
	\section{Related Works}
	
	\paragraph{Offline Reinforcement Learning} Offline RL learns from a collected offline dataset and suffers from the issue of out-of-distribution (OOD). 
	Some prior works propose to constrain the learned policy towards the behavior policy by adding KL-divergence\cite{AW,nair2020awac,wang2020critic,Zhuang2023BehaviorPP}, MSE \cite{PS}, or the regularization of the action selection \cite{BEAR}.
	Some articles suggest that if the collected data is sub-optimal, these approaches usually perform not well \cite{CODAC}. But other works point out that adding a supervised learning term to the policy improvement objective \cite{TD3PlusBC} will also receive high performance by reducing the exploration.
	Another effective way is to learn a conservative Q-function \cite{CQL,Mildly,Implicit}, which assigns a low Q-value for OOD states and then extracts the corresponding greedy policy.
	Moreover, other works propose to use an ensemble model to estimate Q-value \cite{QRDQN} or consider the importance sampling \cite{AW}.
	Such methods have not previously considered the setting of sequential tasks and naive translating them into CORL setting is ineffective, while this paper focuses on sequential tasks and aims to solve the catastrophic forgetting problem during the process of learning a sequence of offline RL tasks.
	
	
	On the other hand, recent works \cite{Chen2021OfflineMA,leeRepresentationBalancingOffline2022} propose to train a dynamic model to predict the values of OOD samples in a supervised-learning way. 
	Such model-based offline RL methods offer great potential for solving the OOD problem, even though the transition model is hardly accurate strictly. The model algorithm is thought to alleviate the OOD problem faced by offline RL and thus improve the robustness of the offline agent.
	Model-based offline RL methods have two major categories:  
	one focuses on measuring the uncertainty of the learned dynamic model \cite{MOPO,MOReL}, and the other considers the pessimistic estimation \cite{COMBO}.
	Different from most of these works using the dynamic model to generate OOD samples when training the agent, in this paper, we utilize the dynamic model to search for the most valuable episodes in the offline dataset for the ER method.
	
	\paragraph{Continual Reinforcement Learning} 
	Offline methods may consider a single-task or multi-task scenario \cite{yu2021conservative,liu2021dara,Liu2023BeyondOS}.
	In contrast, continual learning attempts to learn new tasks after having learned old tasks and get the best possible results on all tasks. 
	Generally, continual learning methods can be classified into three categories \cite{CL}: regularization-based approaches \cite{EWC,SI} add a regularization term to prevent the parameters from far from the value learned from past tasks; 
	modular approaches \cite{PathNet,PackNet} consider fixed partial parameters for a dedicated task; 
	and rehearsal-based methods \cite{GEM,AGEM} train an agent by merging the data of previously learned tasks with that of the current task.
	All three kinds of continual learning methods have been applied for RL tasks \cite{PC,Non-Stationarity,LPGFP,Lipschitz}. 
	Specifically, our work is based on the rehearsal method in an RL setting \cite{ER,SER}. Therefore, we will detail the works involved in this category following.
	
	There are two essential questions to answer in rehearsal-based continual learning. 
	The first is how to choose samples from the whole dataset to store in the replay buffer with a limited size \cite{Coreset}. 
	The most representative samples \cite{ICaRL,GPM} or samples easy to forget \cite{HAL} are usually selected in the replay buffer while random selection has also been used in some works \cite{GDumb,DER}.
	However, these algorithms are designed for image classification and are not applicable to RL. \cite{SER} focuses on replay buffer question sampling in online RL setting.
	The second is how to take advantage of the saved replay samples \cite{Coreset,HAL}.
	In RL, the two most commonly used approaches are BC and perfect memory \cite{Disentangling} in continual RL, where BC is more effective in relieving catastrophic forgetting.
	At present, all these methods are designed for online RL setting.
	Unlike prior works, we consider the offline RL setting in this paper, where catastrophic forgetting and overestimation must be overcome simultaneously.
	
	To the best of our knowledge, this is \emph{the first work} to solve offline RL problems in the continual learning setting. 
	
	
	\section{Problem Formulation and Preliminary}
	
	\paragraph{Continual Offline Reinforcement Learning} In this paper, we investigate CORL, which learns a sequence of RL tasks $\mathcal{T} = \left(T_1, \cdots, T_N\right)$. 
	Each task $T_n$ is described as a Markov Decision Process (MDP) represented by a tuple of $\{\mathcal{S}, \mathcal{A}, P_n, \rho_{0,n}, r_n, \gamma\}$, 
	where $\mathcal{S}$ is the state space, 
	$\mathcal{A}$ is the action space, $P_n:\mathcal{S}\times\mathcal{A}\times\mathcal{S}\rightarrow\left[0,1\right]$ is the transition probability, 
	$\rho_{0,n}:\mathcal{S}$ is the distribution of the initial state, 
	$r_n:\mathcal{S}\times\mathcal{A}\rightarrow\left[-R_\text{max},R_\text{max}\right]$ is the reward function, and $\gamma\in\left[0,1\right)$ is the discounting factor. 
	We assume that sequential tasks have different $P_n, \rho_{0,n}$ and $r_n$, but share the same $\mathcal{S}$, $\mathcal{A}$, and $\gamma$ for simplicity. 
	The return is defined as the sum of discounted future reward $R_{t,n}=\sum_{i=t}^H\gamma^{\left(i-t\right)}r_n\left(s_i,a_i\right)$, where $H$ is the horizon. 
	
	We define a parametric Q-function $Q\left(s,a\right)$ and a parametric policy $\pi\left(a|s\right)$. 
	Q-learning methods train a Q-function by iteratively applying the Bellman operator  $\mathcal{B}^*Q\left(s,a\right)=r\left(s,a\right)+\gamma\mathbb{E}_{s^\prime\sim P\left(s^\prime|s,a\right)}\left(\max_{a^\prime}Q\left(s^\prime,a^\prime\right)\right)$. 
	We also train a transition model for each task $\hat{P}_n\left(s^\prime|s,a\right)$ by using maximum likelihood estimation $\min_{\hat{P}_n}\mathbb{E}_{\left(s,a,s^\prime\right)\sim\mathcal{D}}\left[\log\hat{P}\left(s^\prime|s,a\right)\right]$. 
	We use a multi-head architecture for the policy network $\pi$ to avoid the same-state-different-task problem \cite{Kessler2021SameSD}. 
	In detail, the policy network consists of a feature extractor $\theta_z$ for all tasks and multiple heads $\theta_n, n\in\left[1,N\right]$, where one head is for each task. 
	$\pi_n$ is defined to represent the network with joint parameters $\left[\theta_z,\theta_n \right]$ and $h_n$ is defined to represent the head with parameters $\theta_n$. 
	Our aim is to train sequential tasks all over $\left[T_1, \cdots, T_{N-1}\right]$ sequentially and get a high mean performance and a low forgetting of all the learned tasks without access to data from previous tasks except a small buffer.
	
	In online RL setting, the experiences $e=\left(s,a,s^\prime,r\right)$ can be obtained through environment interaction. 
	However, in offline RL setting, the policy $\pi_n\left(a|s\right)$ can only be learned from a static dataset $\mathcal{D}_n=\left\{e_n^i\right\}, e_n^i=\left(s_n^i,a_n^i,s_n^{\prime i},r_n^i\right)$, which is assumed to be collected by an unknown behavior policy $\pi_n^\beta\left(a|s\right)$.
	
	\paragraph{Experience Replay} 
	ER \cite{ER} is the most widely-used rehearsal-based continual learning method.
	In terms of task $T_n$, the objective of ER is to retain good performance on previous tasks $\left[T_1, \cdots, T_{n-1}\right]$, by using the corresponding replay buffers $\left[B_1,\cdots,B_{n-1}\right]$, which called perform memory.
	Moreover, two additional behavior cloning losses, including the actor cloning loss and the critic cloning loss, are commonly used for previous tasks as follows
	\begin{eqnarray}
	L_\text{actor\_cloning}&:=&\sum\limits_{s,a\in B}\left\|\pi_n\left(s\right)-a\right\|_2^2, \label{actor_replay}\\
	L_\text{critic\_cloning}&:=&\sum\limits_{s,a,Q_\text{replay}\in B}\left\|Q_n\left(a,s\right)-Q_\text{replay}\right\|_2^2, \label{critic_replay}
	\end{eqnarray}
	where $B$ is the replay buffer and $Q_\text{replay}$ means the Q value saved from previous tasks. These two losses are called BC \cite{Disentangling}.
	
	Subsequent work \cite{Disentangling} shows that for a soft actor-critic \cite{Haarnoja2018SoftAA} architecture, the replay loss added on the actor network (Eq. \ref{actor_replay}) performs well, but the loss added on the critic network (Eq. \ref{critic_replay}) poorly effect. 
	Therefore, we only consider the actor cloning loss (Eq. \ref{actor_replay}) in our work.
	
	However, naively integrating the ER with offline RL results in a significant performance drop for CORL problems. 
	To tackle this problem, we propose a novel method OER, which consists of two essential components as follows.

	\section{Offline Experience Replay (OER)}
	In this section, we first elaborate on how to select valuable experiences and build the replay buffer. 
	Then, we describe a novel DBC architecture as our replay model. 
	Finally, we summarize our algorithm and provide the Pseudocode.

	\subsection{Offline Experience Selection Scheme}
	
\begin{figure}
		\centering
		\subfigure[offline learned trajectory]{
			\includegraphics[width=0.45\linewidth]{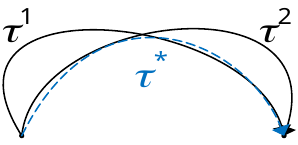}
			\label{MBER1_1}}
		\subfigure[Offline buffer selection]{
			\includegraphics[width=0.45\linewidth]{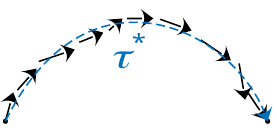}
			\label{MBER1_2}}
		\caption{Distribution shift between the experiences from replay buffer and trajectories from learned policy. (a) $\tau^{1}$ and $\tau^{2}$ represent two trajectories in offline dataset, and $\tau^*$ represents the trajectory generated by the optimal policy. The learned policy can generate better trajectories than original dataset.
			(b) The arrow represents the experiences in offline dataset near the optimal trajectory. Episodes close to the optimal trajectory are more valuably selected.}
		\label{MBER1}
	\end{figure}

	\emph{Novel Distribution Shift Problem:} 
	Facing with sequential tasks, an agent learns the task $T_{n}$ in order after learning $T_{n-1}$, and should prepare for the next learning.
	In terms of task $T_n$, how to select partial valuable data from offline dataset $\mathcal{D}_n$ to build a size-limited replay buffer $B_n$ is critical.
	In online RL setting, rehearsal-based methods commonly utilize a ranking function $\mathcal{R}\left(s_i,a_i\right)=\left|r_i+\gamma\max\limits_{a^\prime}Q\left(s_i^\prime,a^\prime\right)-Q\left(s_i,a_i\right)\right|$ 
	to evaluate the value of trajectories.
	Here, 
	$\mathcal{R}\left(s_i,a_i\right)$ with $\left(s_i,a_i\right)\in\mathcal{D}$ 
	evaluates replay trajectories in terms of $Q$-value accuracy or total accumulative reward \cite{SER}, where the trajectories are collected by the optimal policy $\pi_n^*$ interacting with the environment. 
	On the contrary, in offline RL setting, if we consider the similar method above, the data selection can only be made in the offline dataset $\mathcal{D}_n$ corresponding to the behavior policy $\pi_n^\beta$.
	Therefore, 
	there exists a distribution shift between $\pi_n^\beta$ and $\pi_n^*$, which inevitably affects the performance of the rehearsal. 
	In our selection scheme, we attempt to identify and filter out those offline trajectories in $\mathcal{D}_n$ that are not likely to be generated by $\pi_n^*$. 
	To clarify this point, we give an illustration in Fig.\ref{MBER1}, where those trajectories close to the offline optimal trajectory are selected and stored in the replay buffer $B_n$. The distribution of these trajectories in terms of both $\mathcal{S}$ and $\mathcal{A}$ differs from the offline dataset $\mathcal{D}_n$.
	
	\begin{figure}
		\centering
		\includegraphics[width=0.7\linewidth]{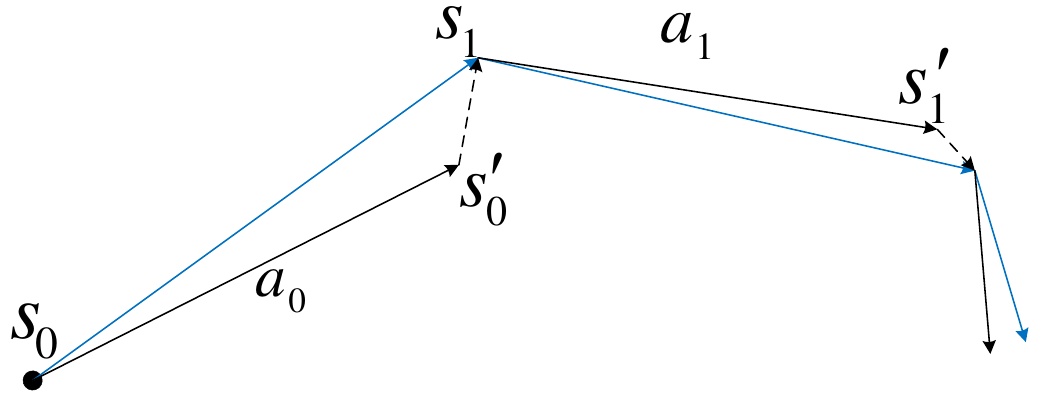}
		\caption{Illustration of our MBES method to fill in $B_n$ for $T_n$. Blue lines represent the experiences in $D_n$, and black lines represent the generated experiences by optimal policy $\pi_n^*$. Starting from $s_0$ and $a_0=\pi_n^*\left(s_0\right)$, the next state is $s^\prime_0=\hat{P}\left(s_0,a_0\right)$. To avoid the accumulated error, we use the most similar state $s_1$ in the offline dataset $\mathcal{D}_n$ instead of the $s^\prime_0$ as the next state.}
		\label{MBER2}
	\end{figure}

	\begin{figure*}[htbp]
		\centering
		\subfigure[offline learned $T^*$]{
			\includegraphics[width=0.4\linewidth]{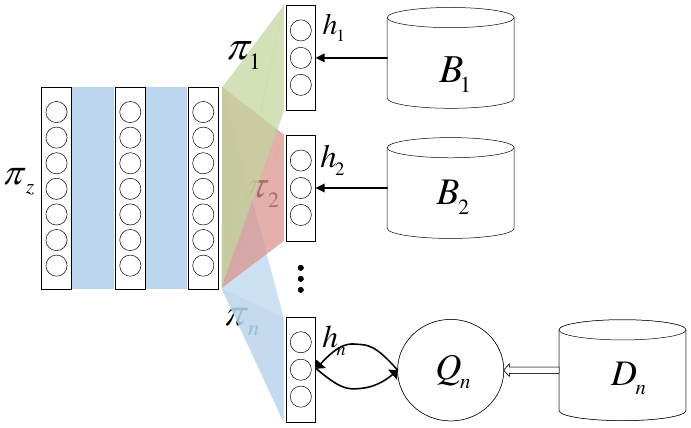}
			\label{DBC_1}}
		\subfigure[Offline buffer selection]{
			\includegraphics[width=0.5\linewidth]{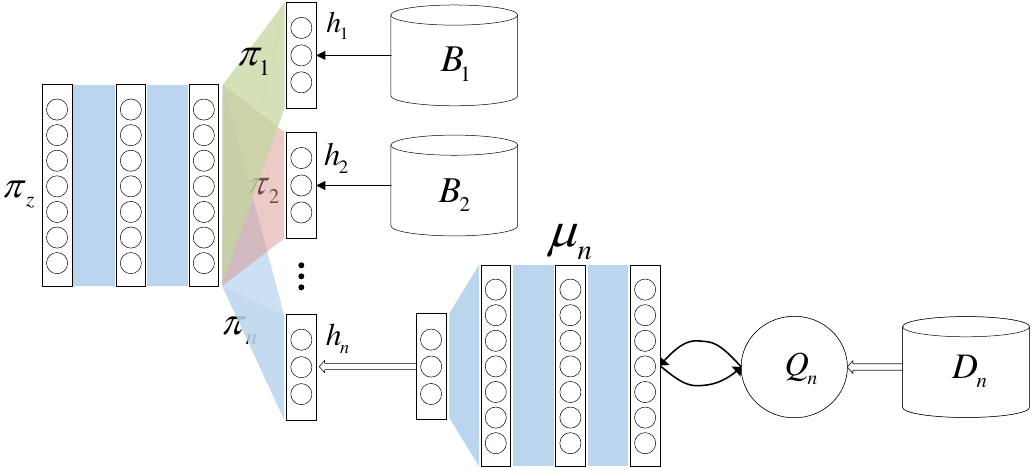}
			\label{DBC_2}}
		\caption{Network architecture of BC and our DBC for experience play, where the policy $\pi_n = \left[\pi_z,h_n\right]$ is for task $n$, $n=1,\cdots,N$. 
			(a) Existing multi-head architecture. Here, the newly-added head $h_n$ for $T_n$ is learned via an actor-critic algorithm with a Q-network $Q_n$, but other heads from $h_1$ to $h_{n-1}$ corresponding to previous tasks are learned by cloning $B_1$ to $B_{n-1}$. 
			(b) Our DBC architecture. We use an independent policy network $\mu_n$ to learn the current task $T_n$ and a newly-added head $h_n$ is used to clone $\mu_n$.}
		\label{DBC}
	\end{figure*}

	\emph{Model-based Experience Selection (MBES):}
	In order to address the new distribution shift question above, we propose a novel MBES scheme. 
	As shown in Fig.\ref{MBER2}, based on the offline trajectories in $\mathcal{D}_n$ collected from $\pi_n^\beta$, MBES aims to generate a new trajectory corresponding to $\pi_n^*$.
	Specifically, MBES considers both the learned optimal policy and a task-specific dynamic model $\hat{P}_n$, where the dynamic model $\hat{P}_n$ is obtained via supervised learning by using $\mathcal{D}_n$. 
	Starting from the $t$th state $s_t$, we recursively sample an action by $\pi_n^*$ and predict the next state $s_t^\prime=\hat{P}_n\left(s_t,\pi_n^*\left(s_t\right)\right)$.
	
	However, recursively sampling actions on predicted states causes a significant accumulation of compounding error after several steps, 
	which hinders the selected trajectories from being revisited later. 
	In order to remove the compounding error, after learning task $n$, starting from the $t$th state $s_t$, instead of directly taking the model output as the next state, a state in $\mathcal{D}_n$ most similar to the model prediction $s_t^\prime$ is selected as the next state $s_{t+1}$ for the pseudo exploitation. 
	Here, we use the $L_2$ metric to measure the similarity as follows
	\begin{equation}\label{s-prime}
	s_{t+1}=\mathop{\mathrm{argmin}}\limits_{s\in\mathcal{D}_n}\text{dist}\left(s,s_t^\prime\right)_{s_t^\prime\sim{\hat{P}_n\left(s_t,\pi_n^*\left(s_t\right)\right)}},
	\end{equation}
	where $\text{dist}$ means a distance between $s$ and $s_t^\prime$. \footnote{Here we choose the L2 distance of the feature from the last hidden layer of the Q network in our experiments. However, we find that a simple L2 distance in the state space $\mathcal{S}$ also works well.}
	According to the model-based offline RL methods \cite{Chen2021OfflineMA,leeRepresentationBalancingOffline2022}, we further introduce the variance of $\hat{P}_n\left(s_t,\pi_n^*\left(s_t\right)\right)$ to determine whether the results of the dynamic model are reliable or not, where we use Eq.\ref{s-prime} for selection only if the variance of the $\hat{P}_n$ is lower than the threshold; otherwise, keeping $\pi_n^\beta\left(s_t\right)$ instead. In our experiments, we specifically use $2\hat{P}_n\left(s_t,a_t\right)$ as the threshold, which has minimal impact.
	
	For a start-up, we sample $s_0$ from $\rho_{0,n}$, and then iteratively carry out.
	In the end, we save the generated trajectory in $B_n$.

	\subsection{Dual Behavior Cloning (DBC)}\label{DBClabel}
	Prior rehearsal-based approaches utilize a multi-head policy network $\pi$ and a Q-network $Q_n$ to learn task $T_n$, as shown in Fig.\ref{DBC} (a).
	During learning, the policy network $\pi$ is to clone the experience data stored in replay buffers $B_1$ to $B_{n-1}$ for all previous tasks $T_1$ to $T_{n-1}$. 
	However, such an architecture suffers from an obvious performance drop when the number of tasks increases, 
	which indicates that the policy $\pi$ trained via BC has the incapability of mastering both previous knowledge from old tasks and new knowledge from the new task.
	
	This performance drop is due to the existing inconsistency between the following two objectives: on the one hand, the multi-head policy $\pi$ is optimized for all current and previous tasks $T_1$ to $T_n$ for action prediction; on the other hand, the policy $\pi$ is also used for updating the current Q-network $Q_n$.
	More specifically, we begin this analysis from Q-learning \cite{PlayAtari,HumanLevel} with the Bellman equation of $Q\left(s,a\right)=r\left(s,a\right)+\gamma*\max_{a^\prime}Q\left(s^\prime,a^\prime\right)$. 
	In a continuous action space, as the maximization operator above cannot be realized, a parameterized actor network $\mu\left(s\right)$ is commonly used instead to take the optimal action corresponding to maximal Q value. 
	$\mu$ and $\pi$ can be considered equivalently in a single-task setting. 
	However, for continual learning, $\pi$ is constrained to clone all previous tasks from $T_1$ to $T_{n-1}$ while $\mu$ depends only on the current task $T_n$. 
	Consequently, it is difficult for the policy $\pi$ to take action corresponding to the maximum future reward for $s^\prime$ in task $T_n$.
	
	Based on such analysis, we propose a novel DBC scheme to solve the inconsistency mentioned above, and the schematic diagram of our proposed architecture is given in Fig. \ref{DBC_2}. 
	In comparison to existing multi-head network architecture in Fig. \ref{DBC_1} \cite{ER}, 
	we propose an extra policy network $\mu_n$ in order to learn the optimal state-action mapping for $T_n$.
	Specifically, when learning task $T_n$, we first obtain $\mu_n$ and $Q_n$ by using an offline RL algorithm. 
	Then, in the rehearsal phase, the continual learning policy $\pi$ is required to clone the previous experiences from $B_1$ to $B_{n-1}$ and meanwhile be close to $\mu_n$. Thus, the corresponding loss $L_\pi$ can be written as follows 
	\begin{eqnarray}\label{P}
	L_\pi&=&\mathbb{E}_{\left(s,a,s'\right)\sim\mathcal{D}_n}\left(\pi_n(s, a)-\mu_n\left(s,a\right)\right)^2\nonumber\\&+&\lambda_r\frac{1}{n}\sum\limits_{j=1}^{n-1}\mathbb{E}_{\left(s,a\right)\sim\mathcal{B}_{j}}\left(\pi_{j}\left(s\right)-a\right)^2,
	\end{eqnarray}
	where $\lambda_r$ is a coefficient for the BC item.
	It is worth mentioning that the continual learning policy $\pi$ attempts to clone the behavior of both $\mu_n$ and the previous experiences from $B_1$ to $B_{n-1}$ simultaneously. Therefore, we name our scheme DBC.
	
	\subsection{Algorithm Summary}
	When considering a new task $T_n$, we first utilize DBC to learn the two policy networks $\pi$, $\mu_n$, and the dynamic model $\hat{P}_n$ until convergence. 
	Then, the learned policy $\pi$ and the dynamic model $\hat{P}_n$ are used in MBES to select valuable data for $B_n$.
	To summarize, the overall process of OER, including DBC and MBES, is given in Algorithm \ref{algo}.
	\begin{algorithm}[h]
		\caption{Our proposed Method OER}
		\label{algo}
		\begin{algorithmic}[1]
			\REQUIRE Number of tasks $N$; initiate the policy $\pi$.
			\FOR{Tasks $T_n$ in $\left[1,\cdots,N\right]$}
			\STATE Get offline dataset $\mathcal{D}_n$; Initiate the replay buffer $B_n=\emptyset$; Initiate new head $h_n$ for $\pi$; Initiate $\mu_n$, $Q_n$ and $\hat{P}_n$.
			\WHILE{Not Convergence}
			\STATE Update $\mu_n$ and $Q_n$ via offline learning method.
			\STATE Update $\pi$ via Eqn. \ref{P} to clone $\mu_n$ and $B_0$ to $B_{n-1}$.
			\STATE Update the dynamic model $\hat{P}_n$.
			\ENDWHILE
			\STATE Sample initial state $s_0$ from $\rho_{0,n}$, and $s_t\leftarrow s_0$.
			\WHILE{$B_n$ is not full}
			\IF{$s_t$ is not terminal state}
			\STATE Select $s_{t+1}$ from $\mathcal{D}_n$ by $\pi_n$ and $\hat{P}_n$ via Eqn. \ref{s-prime}.
			\ELSE
			\STATE Sample $s_{t+1}$ from $\rho_{0,n}$.
			\ENDIF
			\STATE Add $s_{t+1}$ into $B_n$, and $s_t\leftarrow s_{t+1}$.
			\ENDWHILE
			\ENDFOR
			\ENSURE $\pi$.
		\end{algorithmic}
	\end{algorithm}
	
	\section{Implementation Details}
	We model the Q-function and policy network as a multi-layer perceptron (MLP) with two hidden layers of 128 neurons, each with ReLU non-linearity based on \cite{d3rlpy}. 
	We use an ensemble dynamic model containing five individual models, which is also modeled as an MLP, the same as the Q-function and the policy network.
	Any offline RL method is compatible with our architecture. 
	We choose TD3+BC \cite{TD3PlusBC} as the backbone algorithm because of its simple structure to demonstrate that our algorithm does not depend on a specific offline algorithm. 
	Further, we use Adam \cite{Adam} with a learning rate of $0.001$ to update both the Q-function and the dynamic model and $0.003$ to update both the policy network $\pi$ and $\mu_n$.
	Then, for each task, we train $30,000$ steps and switch to the next. 
	We find that initializing $\mu_n$ with $\pi_{n-1}$ and learning $\pi$ and $\mu_n$ simultaneously work well from the experience. Also, learning $\pi$ and $\mu_n$ together will reduce the scope of the gradient and avoid the jumping change to ensure stable learning and relieve catastrophic forgetting.
	The result is calculated via five repeated simulations with different numbers of seeds.

	\section{Experiments}
	Extensive experiments are conducted to demonstrate the effectiveness of our proposed scheme and test whether we can keep both stability and plasticity at the same time when learning sequential offline RL tasks. We evaluate the performance of MBES and DBC separately to test the performance of each approach. 
	
	\subsection{Baselines and Datasets}
	
	\paragraph{Baselines}
	\emph{On one hand}, in order to evaluate the MBES, we consider six replay buffer selection approaches, where four from \cite{SER} are given as follows.
	\begin{itemize}
		\item Surprise \cite{SER}: store trajectories with maximum mean TD error: $\min_{\tau\in D_i}\mathbb{E}_{s,a,s_{t+1}\in\tau}\left\|\mathcal{B}^*Q\left(s,a\right)-Q\left(s,a\right)\right\|_2^2$.
		\item Reward \cite{SER}: store trajectories with maximum mean reward: $\max_{\tau\in D_i}R_{t,n}$.
		\item Random \cite{SER}: randomly select samples in dataset.
		\item Coverage \cite{SER}: store those samples to maximize coverage of the state space, or ensure the sparsity of selected states: $\min_{s\in D_i}\left|\mathcal{N}_i\right|; \mathcal{N}_i=\left\{s' \st\text{dist}\left(s'-s\right)<d\right\}$.
	\end{itemize}
	Considering that these baselines are designed for online RL and it is not fair enough to use them only as the baselines, we have designed two algorithms for comparison that are applicable to offline RL, based on the idea of \cite{SER}.
	\begin{itemize}
		\item Match: select samples in the offline dataset most consistent with the learned policy. Trajectories chosen in this way are most similar to the learned policy in the action space, but may not match in the state space: $\min_{\tau\in D_i}\mathbb{E}_{s,a\in\tau}\left\|a-\pi_i^*\left(s\right)\right\|_2^2$.
		\item Model: Given that we used a Model-Based approach to filter our data, we also used it as a criterion for whether the trajectories matched the transfer probabilities: $\min_{\tau\in D_i}\mathbb{E}_{s,a\in\tau}\left\|\hat{P}_i\left(s,a\right)-P_i\left(s,a\right)\right\|_2^2$. This metric is used to demonstrate that the introduction of the Model-Based approach alone does not improve the performance of the CORL algorithm.
	\end{itemize}
	
	\emph{On the other hand}, in order to evaluate the DBC, we consider five widely-used continual learning methods, where three methods need to use replay buffers.
	\begin{itemize}
		\item BC \cite{ER}: a basic rehearsal-based continual method adding a behavior clone term in the loss function of the policy network.
		\item Gradient episodic memory (GEM) \cite{GEM}: a method using an episodic memory of parameter gradients to limit the policy update.
		\item Averaged gradient episodic memory (AGEM) \cite{AGEM}: a method based on GEM that only uses a batch of gradients to limit the policy update.
	\end{itemize}
	In addition, the following two regularization-based methods are rehearsal-free so that they are independent of experience selection methods. 
	\begin{itemize}
		\item Elastic weight consolidation (EWC) \cite{EWC}: constrain the changes to critical parameters through the Fisher information matrix.
		\item Synaptic intelligence (SI) \cite{SI}: constrain the changes after each step of optimization.
	\end{itemize}
	
	We also show the performance on multi-task as a reference. The multi-task learning setting does not suffer from the catastrophic forgetting problem and can be seen as superior.
	
	\paragraph{Offline Sequential Datasets}
	We consider three sets of tasks from widely-used continuous control offline meta-RL library\footnote{Most current continual RL methods perform not well on complex benchmarks such as \cite{ContinualWorld}, so it is not suitable for evaluating our method \cite{metaORL} and we do not consider it in this paper.} as in \cite{metaORL}:
	\begin{itemize}
		\item Ant-2D Direction (Ant-Dir): train a simulated ant with 8 articulated joints to run in a 2D direction;
		\item Walker-2D Params (Walker-Par): train a simulated agent to move forward, where different tasks have different parameters. Specifically, different tasks require the agent to move at different speeds;
		\item Half-Cheetah Velocity (Cheetah-Vel): train a cheetah to run at a random velocity.
	\end{itemize} 
	For each set of tasks, we randomly sample five tasks to form sequential tasks 
	$T_1$ to $T_5$.
	
	To consider different data quality as \cite{d4rl}, we train a soft actor-critic to collect two benchmarks \cite{SAC} for each task $T_n$, $n=1,\cdots,5$:
	1) Medium (M) with trajectories from medium-quality policy, and 2) Medium-Random (M-R) including trajectories from both medium-quality policy and trajectories randomly sampled.

	\paragraph{Metrics}
	Following \cite{KernelCL}, we adopt the average performance (PER) and the backward transfer (BWT) as evaluation metrics,
	\begin{equation}
	\text{PER}=\frac{1}{N}\sum\limits_{n=1}^Na_{N,n},\ \text{BWT}=\frac{1}{N-1}\sum\limits_{n=1}^{N-1}a_{n,n}-a_{N,n},
	\end{equation}
	where $a_{i,j}$ means the final cumulative rewards of task $j$ after learning task $i$. For PER, higher is better; for BWT, lower is better.
	These two metrics show the performance of learning new tasks while alleviating the catastrophic forgetting problem.
	
	\begin{table*}[ht]
		\centering
		\begin{tabular}{c|c|c|c|c|c|c|c}
			\hline
			\multirow{2}{*}{Benchmark}&\multirow{2}{*}{Methods}&\multicolumn{2}{|c|}{Ant-Dir}&\multicolumn{2}{|c|}{Walker-Par}&\multicolumn{2}{|c}{Cheetah-Vel}\\
			\cline{3-8}
			&&PER&BWT&PER&BWT&PER&BWT\\
			\hline
			\multirow{9}{*}{M-R}&MultiTask&1387.54&-&1630.90&-&-112.49&-\\
			&Coverage+DBC&677.72&727.40&231.17&1428.14&-404.33&293.42\\
			&Match+DBC&776.85&432.74&120.96&1079.19&-121.30&26.50\\
			&Supervise+DBC&903.12&175.17&196.07&1063.35&-233.05&82.60\\
			&Reward+DBC&893.63&36.10&554.05&1087.96&-242.50&102.75\\
			&Model+DBC&1156.43&65.66&194.31&1214.32&-141.20&57.90\\
			&Random+DBC&845.82&126.56&614.71&979.90&-104.16&36.24\\
			&OER&\textbf{1316.46}&\textbf{119.71}&\textbf{1270.62}&\textbf{550.18}&\textbf{-76.61}&\textbf{16.54}\\
			\hline
			\multirow{8}{*}{M}&MultiTask&1357.20&-&1751.68&-&-115.30&-\\
			&Coverage+DBC&842.46&599.74&361.55&1342.87&-424.87&229.89\\
			&Match+DBC&841.88&549.98&886.28&678.55&-196.83&88.87\\
			&Supervise+DBC&1049.84&347.88&1020.57&666.15&-219.78&56.99\\
			&Reward+DBC&1125.44&269.76&891.55&790.07&-222.83&40.15\\
			&Model+DBC&1147.05&253.49&872.94&807.33&-184.83&24.22\\
			&Random+DBC&1189.17&179.75&1102.89&616.52&-150.39&30.18\\
			&OER&\textbf{1215.58}&\textbf{176.85}&\textbf{1192.59}&\textbf{518.20}&\textbf{-148.18}&\textbf{16.36}\\
			\hline
		\end{tabular}
		\caption{Performance of our OER and baselines to verify the effectiveness of MBES, where M-R and M are included. We can observe that our method OER has the highest PER and lowest BWT in all cases.}
		\label{Performance MBES}
	\end{table*}
	\begin{table*}[ht]
		\centering
		\begin{tabular}{c|c|c|c|c|c|c|c}
			\hline
			\multirow{2}{*}{Benchmark}&\multirow{2}{*}{Methods}&\multicolumn{2}{|c|}{Ant-Dir}&\multicolumn{2}{|c|}{Walker-Par}&\multicolumn{2}{|c}{Cheetah-Vel}\\
			\cline{3-8}
			&&PER&BWT&PER&BWT&PER&BWT\\
			\hline
			\multirow{6}{*}{M-R}
			&MBES+SI&747.95&643.89&124.04&1460.08&-437.49&316.84\\
			&MBES+EWC&655.21&726.37&110.53&1589.62&-568.07&506.19\\
			&MBES+GEM&748.90&643.06&114.07&1477.87&-445.39&389.10\\
			&MBES+AGEM&722.41&687.97&62.27&1628.53&-546.15&419.61\\
			&MBES+BC&407.94&874.80&46.75&860.17&-645.79&21.92\\
			&OER&\textbf{1316.46}&\textbf{119.71}&\textbf{1270.62}&\textbf{550.18}&\textbf{-76.61}&\textbf{16.54}\\
			\hline
		\end{tabular}
		\caption{Performance of our OER and baselines to verify the effectiveness of DBC, where M-R is included. The result of M can be found in the Supplementary Material. We can observe that our method OER has the highest PER and lowest BWT in all cases.}
		\label{Performance DBC}
	\end{table*}

	\subsection{Overall Results}
\begin{figure*}[ht]
		\centering
		\subfigure[Match M]{
			\includegraphics[width=0.23\linewidth]{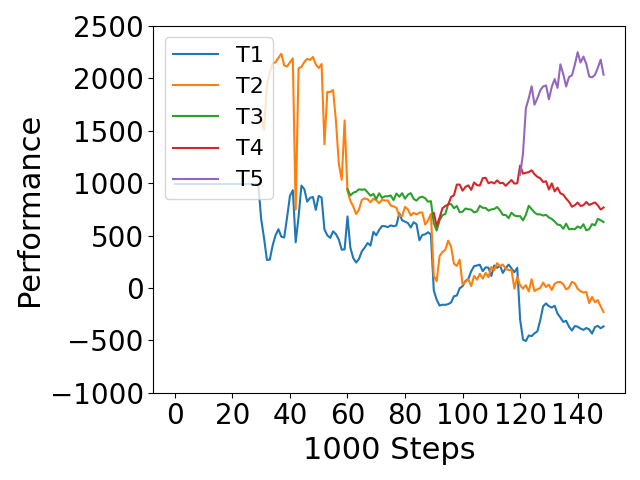}
			\label{maxmatch}}
		\subfigure[Match M-R]{
			\includegraphics[width=0.23\linewidth]{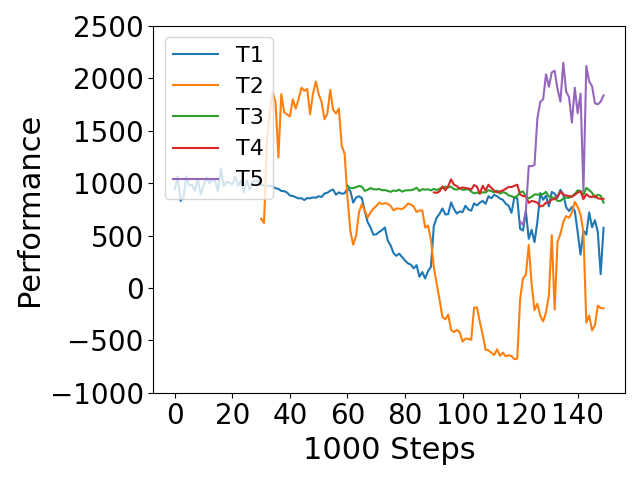}
			\label{maxmatchmr}}
		\subfigure[Random M]{
			\includegraphics[width=0.23\linewidth]{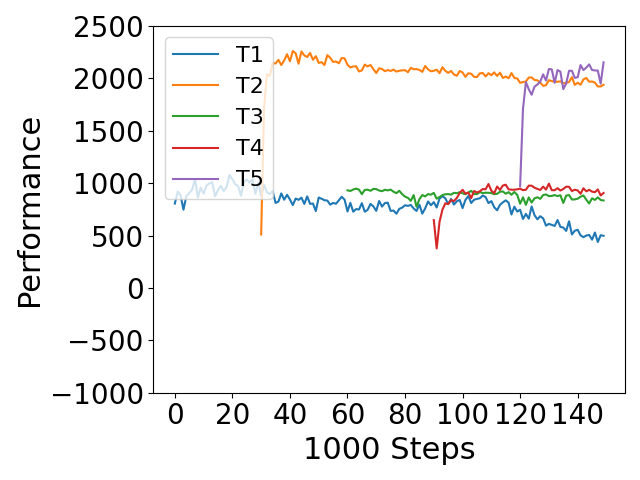}
			\label{random}}
		\subfigure[Random M-R]{
			\includegraphics[width=0.23\linewidth]{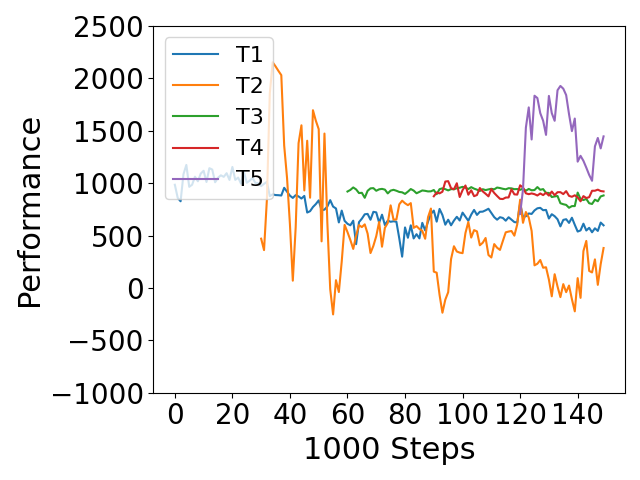}
			\label{randommr}}
		\subfigure[EWC M]{
			\includegraphics[width=0.23\linewidth]{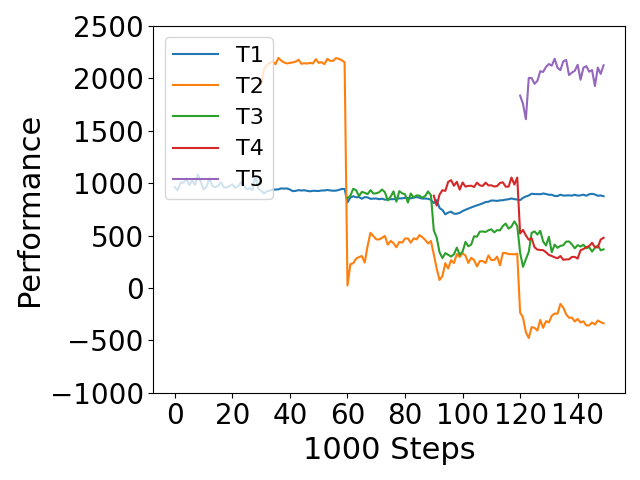}
			\label{ewc}}
		\subfigure[BC M]{
			\includegraphics[width=0.23\linewidth]{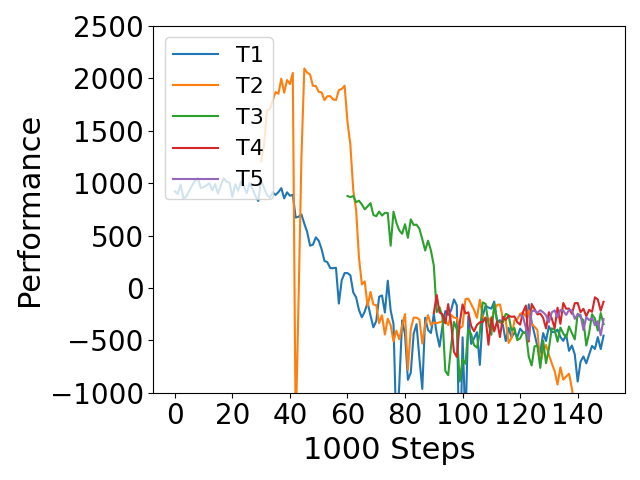}
			\label{bc}}
		\subfigure[OER M]{
			\includegraphics[width=0.23\linewidth]{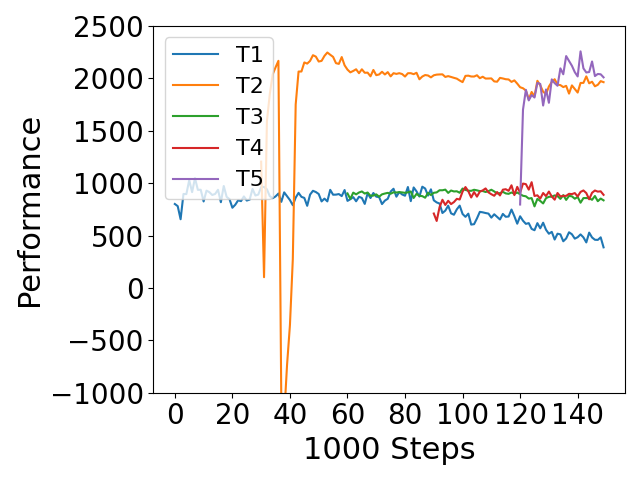}
			\label{dc}}
		\subfigure[OER MR]{
			\includegraphics[width=0.23\linewidth]{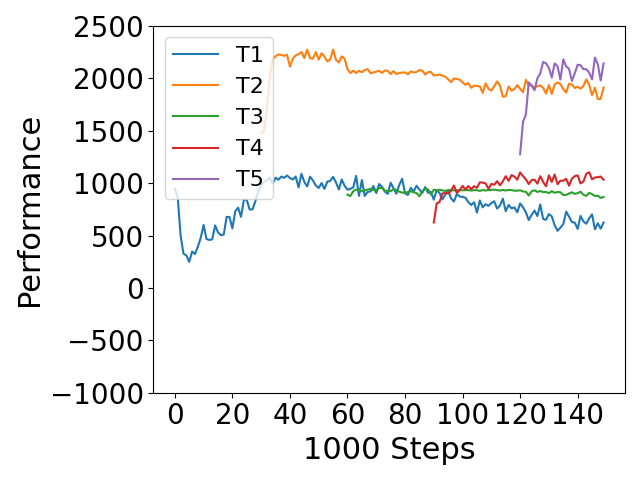}
			\label{dcmr}}
		\caption{Process of learning five sequential tasks, where our OER is compared with Match (M-R and M), Random (M-R and M), EWC (M) and BC (M). More results are given in Supplementary Material. Every 30000 steps on one task, we switch to the next task.}
		\label{Diff}
	\end{figure*}

	\paragraph{Evaluate MBES:} 
	Firstly, our method OER is compared with eleven baselines on two kinds of qualities M-R and M. 
	Since our OER comprises MBES and DBC, six experience selection approaches are added with DBC for a fair comparison.
	The overall results are reported in Table \ref{Performance MBES}, in terms of PER and BWT on three sequential tasks. 
	From Table \ref{Performance MBES}, we draw the following conclusions:
	1) Our OER method outperforms all baselines for all cases, indicating the effectiveness of our MBES scheme;
	2) Our OER method demonstrates greater superiority on M-R dataset than on M dataset, indicating that M-R dataset has a larger distribution shift than M dataset, and MBES addresses such shift;
	3) Random+DBC performs better than the other five baselines because these five experience selection schemes are dedicated for online but not offline scenarios.
	Furthermore, Fig. \ref{Diff} shows the learning process of Ant-Dir on five sequential tasks.
	From Fig.\ref{maxmatch} - \ref{randommr}, Fig.\ref{dc} - \ref{dcmr} and Supplementary Material, we can observe that compared with baselines, our OER demonstrates less performance drop with the increment of tasks, indicating that OER can better solve the catastrophic forgetting.

	\paragraph{Evaluate DBC:} 
	Secondly, our method OER is compared with five baselines. 
	Similarly, continual learning approaches are added with MBES for a fair comparison, and the overall performance is reported in Table \ref{Performance DBC} and the Supplementary Material. 
	From Table \ref{Performance DBC} and the Supplementary Material, we draw the following conclusions:
	1) Our OER method outperforms all baselines, indicating the effectiveness of our DBC scheme;
	2) Four continual learning methods perform not well due to the forgetting problem;
	3) MBES+BC performs the worst due to the inconsistency of two policies $\pi$ and $\mu_n$ in Section \ref{DBClabel}.
	From Fig.\ref{ewc} - \ref{dcmr} and Supplementary Material, we can observe that our OER can learn new tasks and remember old tasks well; other continual learning methods can only learn new tasks but forget old tasks, while BC cannot even learn new tasks.
	
	\begin{table*}[ht]
		\centering
		\begin{tabular}{c|c|c|c|c|c|c|c}
			\hline
			\multicolumn{2}{c|}{\multirow{2}{*}{Method}}&\multicolumn{2}{|c|}{Ant-Dir Medium}&\multicolumn{2}{|c|}{Walker-Par Medium}&\multicolumn{2}{|c}{Cheetah-Vel Medium}\\
			\cline{3-8}
			\multicolumn{2}{c|}{}&PER&BWT&PER&BWT&PER&BWT\\
			\hline
			\multirow{2}{*}{Reward}
			&BC&-308.10&1049.32&124.54&1102.85&-706.31&343.23\\
			&DBC&1310.45&78.89&1528.63&157.11&-207.76&39.82\\
			\hline
			\multirow{2}{*}{Random}
			&BC&-295.06&997.96&115.14&1199.63&-750.20&295.36\\
			&DBC&\textbf{1374.54}&32.01&1565.50&\textbf{45.42}&-143.10&56.16\\
			\hline
			\multirow{2}{*}{MBES}
			&BC&-338.03&1164.01&588.26&958.38&-582.93&199.81\\
			&DBC&1367.07&\textbf{6.32}&\textbf{1578.30}&78.72&\textbf{-141.17}&\textbf{21.18}\\
			\hline
		\end{tabular}
		\caption{Performance of baselines Reward, Random and our OER, where the replay buffer capacity is as 10,000 samples for BC and DBC.}
		\label{MemorySizeChoose}
	\end{table*}

	
	\subsection{Parameter Analysis}
	
	\paragraph{Size of Buffer $B_n$} 
	In rehearsal-based continual learning, the size of replay buffer is a key factor. 
	In our CORL setting, the size of buffer $B_n$ is selected as $1 \ \text{K}$ for all $n$, by considering both storage space and forgetting issues. 
	With the increase of buffer size, we need more storage space but have less forgetting issue.
	In order to quantify such analysis, we consider a buffer size $10 \ \text{K}$ for OER and two baselines, and the results are listed in Table \ref{MemorySizeChoose}.
	From Table \ref{MemorySizeChoose}, we can observe that
	1) With the increase of buffer size, OER and two baselines achieve better performance as expected.
	2) Our DBC method is still much better than BC, indicating that solving the inconsistency is significant;
	3) With larger storage space, the baseline Random performs similar as MBES, because in this case forgetting issue gets much smaller and the experience selection becomes not important.

	\paragraph{Replay Coefficient $\lambda_r$} 
	Another key factor is the coefficient $\lambda_r$ in Eq. \ref{P}, where $\lambda_r$ is used to balance the anti-forgetting BC item and the new-policy constraint item. 
	In our CORL setting, we select $\lambda_r$ as $1$, which is also the general choice in ER-based approaches \cite{ER,SER}, and good performance has been achieved, as mentioned above.
	We analyze different values of $\lambda_r$ and show the corresponding performance of our OER and baselines in Table \ref{ReplayAlpha}, where $\lambda_r$ is selected as $0.3$, $1$ and $3$, respectively.
	From Table \ref{ReplayAlpha}, we can observe that
	with a larger $\lambda_r$, the forgetting issue gradually reduces, but it gets looser that the learning policy $\pi$ clones $\mu_n$ in Eq. \ref{P}, and vice versa.
	As a result, we achieve the best performance when $\lambda_r=1$. 
	This is why we use $\lambda_r=1$ for all experiments in this paper.
	
	\section{Conclusion}
	In this work, we formulate a new CORL setting and present a new method OER, primarily including two key components: MBES and DBC. 
	We point out a new distribution bias problem and training instability unique to the new CORL setting.
	Specifically, to solve a novel distribution shift problem in CORL, we propose a novel MBES scheme to select valuable experiences from the offline dataset to build the replay buffer.
	Moreover, in order to address the inconsistency issue between learning the new task and cloning old tasks, we propose a novel DBC scheme.
	Experiments and analysis show that OER outperforms SOTA baselines on various continuous control tasks.

	\begin{table}
		\centering
		\begin{tabular}{c|c|c|c|c|c}
			\hline
			\multirow{2}{*}{Method}&\multirow{2}{*}{$\lambda_r$}&\multicolumn{2}{|c|}{Ant-Dir M}&\multicolumn{2}{|c}{Walker-Par M}\\
			\cline{3-6}
			&&PER&BWT&PER&BWT\\
			\hline
			\multirow{3}{*}{\makecell[c]{Supervise\\+DBC}}&0.3&980.25&537.92&1003.92&688.19\\
			&1&1049.84&347.88&1020.57&666.15\\
			&3&984.62&201.70&944.71&608.82\\
			\hline
			\multirow{3}{*}{\makecell[c]{Random\\+DBC}}&0.3&1168.82&184.57&1052.13&647.71\\
			&1&1189.17&179.75&1102.89&616.52\\
			&3&952.46&168.81&1157.30&330.47\\
			\hline
			\multirow{3}{*}{OER}&0.3&940.38&431.13&1049.55&606.18\\
			&1&1215.58&176.85&1192.59&518.20\\
			&3&1128.66&170.32&1051.63&507.05\\
			\hline
		\end{tabular}
		\caption{Performance of our OER method with different coefficient $\lambda_r$. High $\lambda_r$ indicates more on replaying previous tasks.}
		\label{ReplayAlpha}
	\end{table}


	\bibliographystyle{ecai}
	\bibliography{ijcai22}
	
	\pagebreak
	\begin{center}
		\textbf{\large Supplemental Materials: OER: Offline Experience Replay for Continual Offline Reinforcement Learning}
	\end{center}
	\setcounter{equation}{0}
	\setcounter{figure}{0}
	\setcounter{table}{0}
	\setcounter{page}{1}
	\makeatletter
	\renewcommand{\theequation}{S\arabic{equation}}
	\renewcommand{\thefigure}{S\arabic{figure}}
	\newcommand{\bibnumfmt}[1]{[S#1]}
	\newcommand{\citenumfont}[1]{S#1}

	\section{Section 1}
	\section{Additional Results}
	Here we give out the learning process of different methods of Ant-Dir dataset in Fig.A1.
	\begin{figure*}[t]
		\centering
		\subfigure[Coverage M]{
			\includegraphics[width=0.23\linewidth]{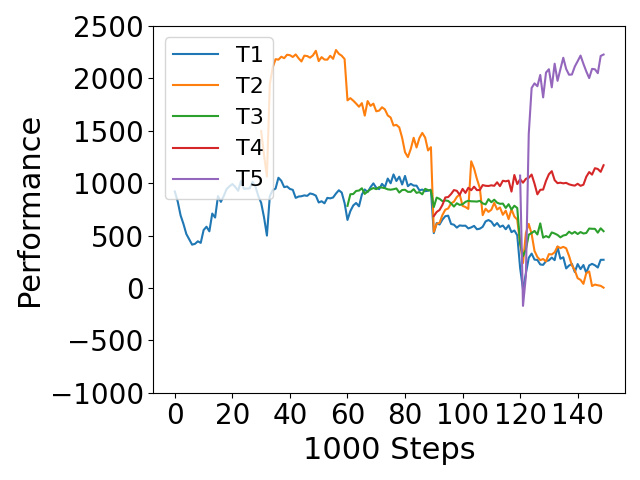}}
		\subfigure[Coverage M-R]{
			\includegraphics[width=0.23\linewidth]{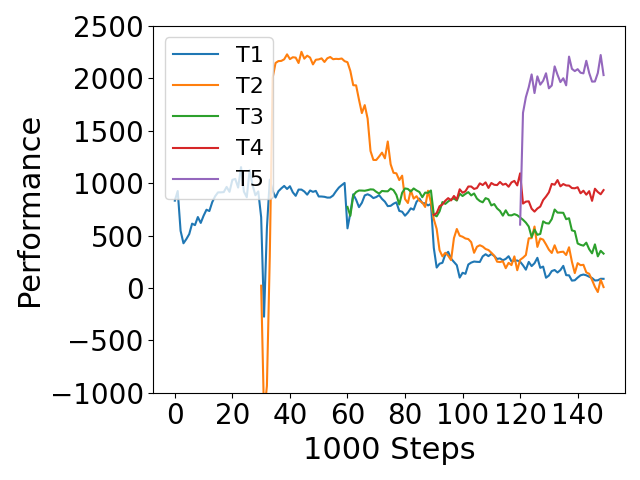}}
		\subfigure[Match M]{
			\includegraphics[width=0.23\linewidth]{./supplementary_results/match_ant_dir_M.png}}
		\subfigure[Match M-R]{
			\includegraphics[width=0.23\linewidth]{./supplementary_results/match_ant_dir_MR.png}}
		\subfigure[Model M]{
			\includegraphics[width=0.23\linewidth]{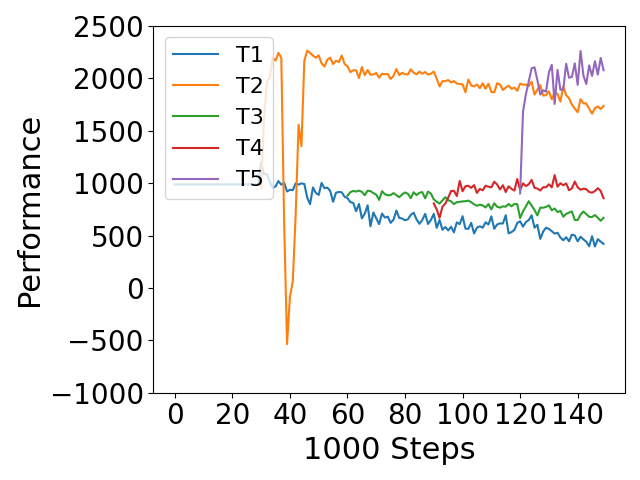}}
		\subfigure[Model M-R]{
			\includegraphics[width=0.23\linewidth]{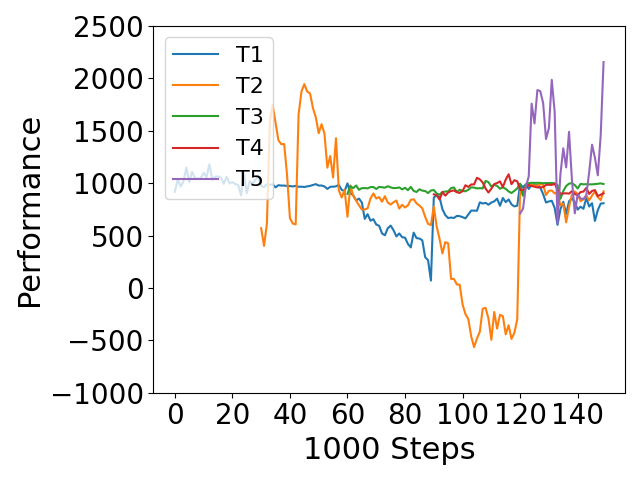}}
		\subfigure[Reward M]{
			\includegraphics[width=0.23\linewidth]{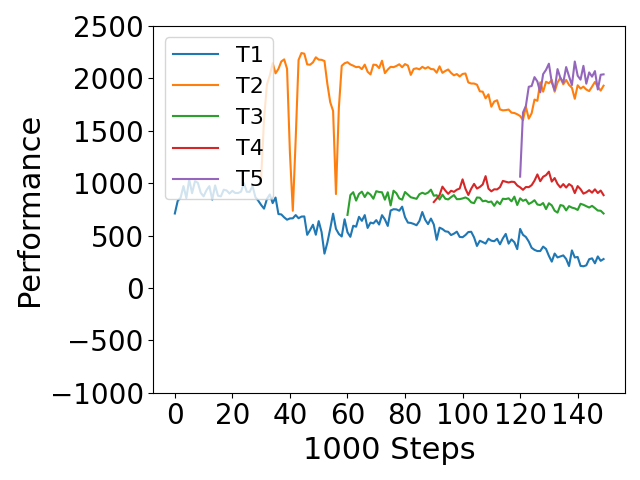}}
		\subfigure[Reward M-R]{
			\includegraphics[width=0.23\linewidth]{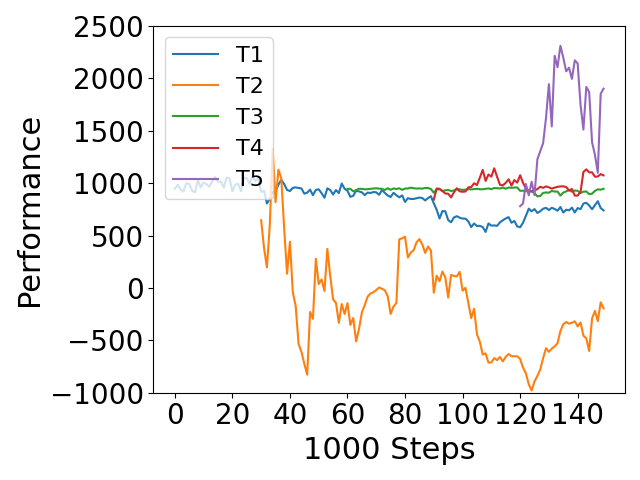}}
		\subfigure[Supervise M]{
			\includegraphics[width=0.23\linewidth]{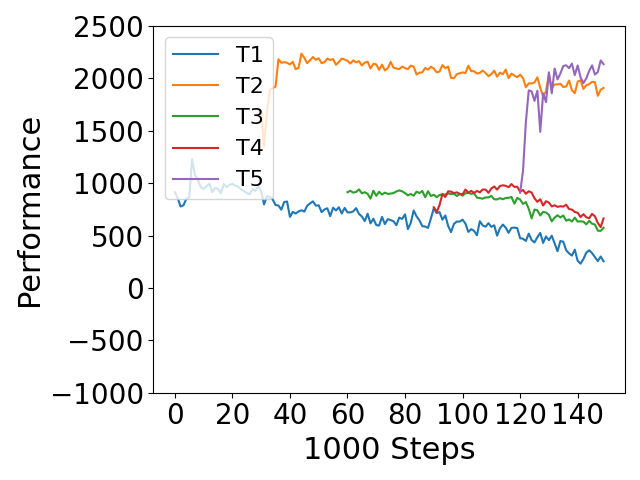}}
		\subfigure[Supervise M-R]{
			\includegraphics[width=0.23\linewidth]{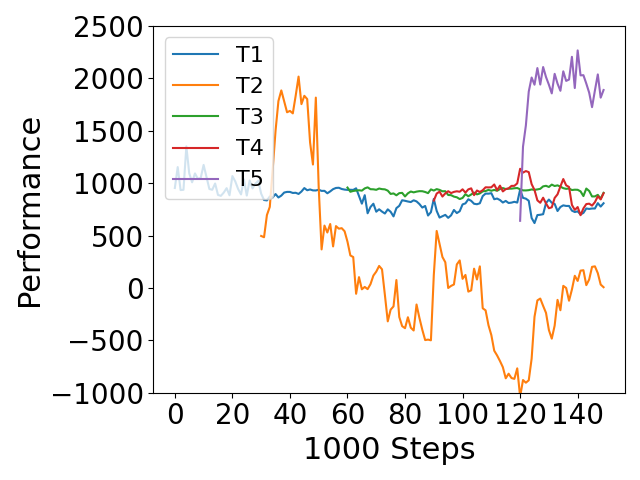}}
		\subfigure[Random M]{
			\includegraphics[width=0.23\linewidth]{./supplementary_results/random_transition_ant_dir_M.png}}
		\subfigure[Random M-R]{
			\includegraphics[width=0.23\linewidth]{./supplementary_results/random_transition_ant_dir_MR.png}}
		\subfigure[GEM M]{
			\includegraphics[width=0.23\linewidth]{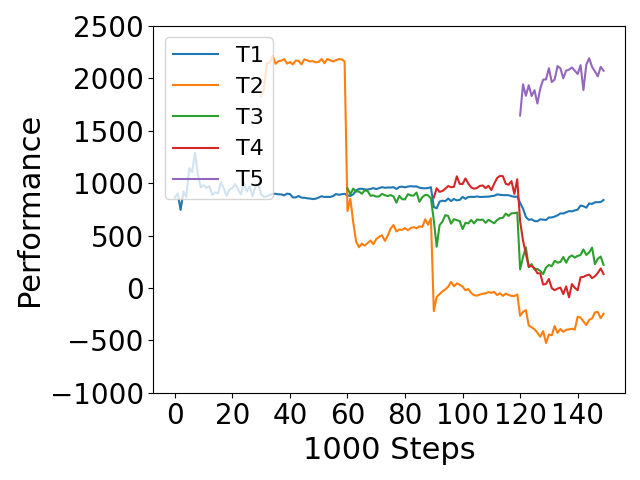}}
		\subfigure[GEM M-R]{
			\includegraphics[width=0.23\linewidth]{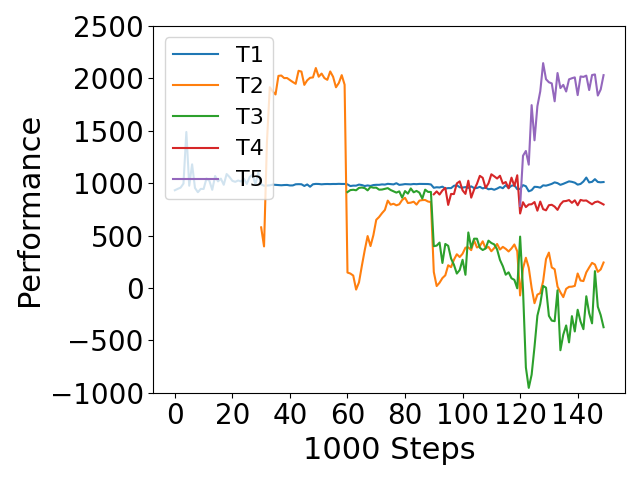}}
		\subfigure[AGEM M]{
			\includegraphics[width=0.23\linewidth]{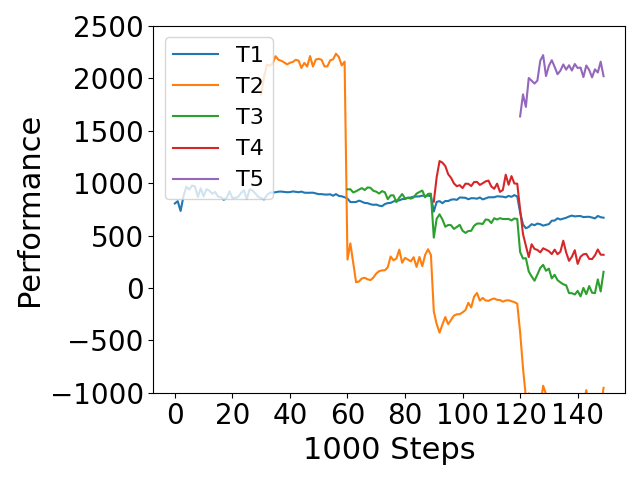}}
		\subfigure[AGEM M-R]{
			\includegraphics[width=0.23\linewidth]{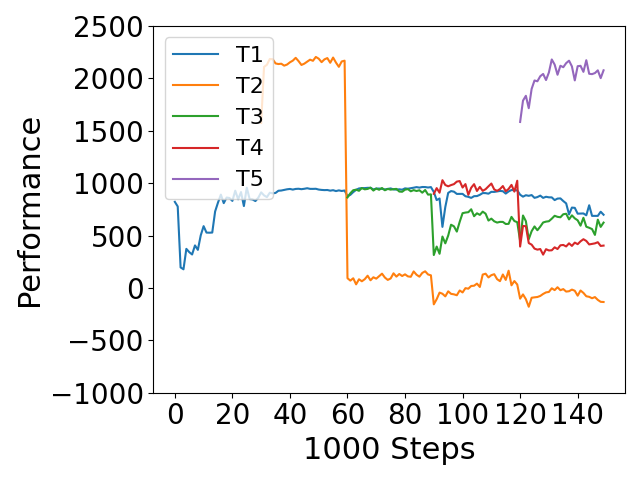}}
		\subfigure[EWC M]{
			\includegraphics[width=0.23\linewidth]{./supplementary_results/ewc_reward_ant_dir_M_bc.png}}
		\subfigure[EWC M-R]{
			\includegraphics[width=0.23\linewidth]{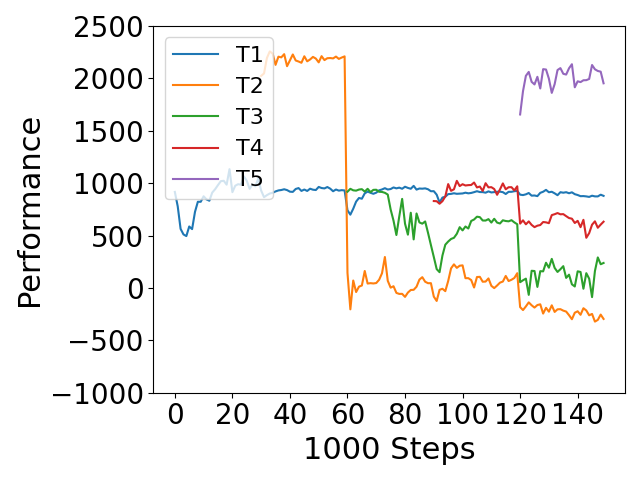}}
		\subfigure[SI M]{
			\includegraphics[width=0.23\linewidth]{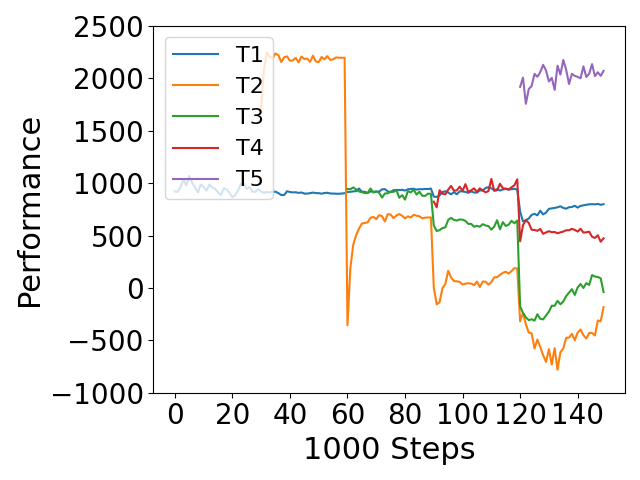}}
		\subfigure[SI M-R]{
			\includegraphics[width=0.23\linewidth]{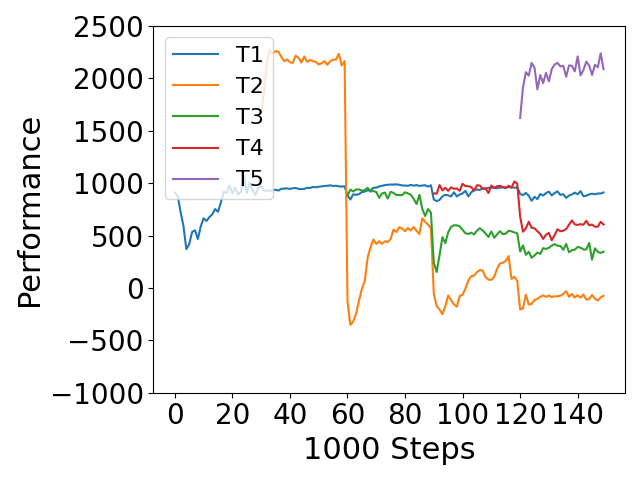}}
		\subfigure[BC M]{
			\includegraphics[width=0.23\linewidth]{./supplementary_results/oer_ant_dir_M_bc.png}}
		\subfigure[BC M-R]{
			\includegraphics[width=0.23\linewidth]{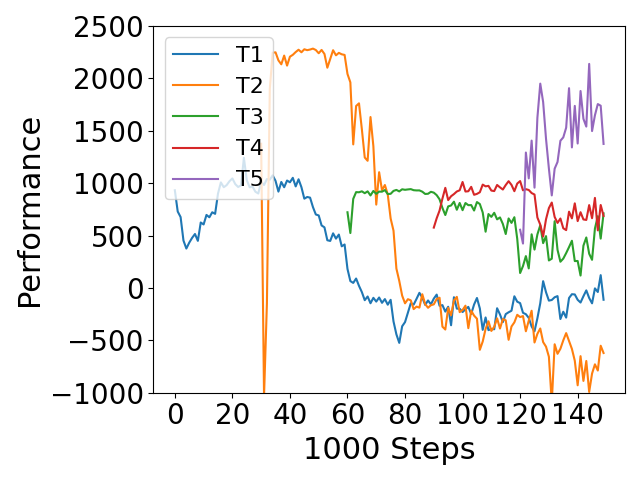}}
		\subfigure[OER M]{
			\includegraphics[width=0.23\linewidth]{./supplementary_results/oer_ant_dir_M.png}}
		\subfigure[OER M-R]{
			\includegraphics[width=0.23\linewidth]{./supplementary_results/oer_ant_dir_MR.png}}

		\begin{center}
			
			Figure 1. Process of learning five sequential tasks, where some of the results have been given in the main body.
			
		\end{center}
		\label{Diff}
	\end{figure*}
\end{document}